\documentclass[english,12pt]{article}
\usepackage{babel}
\title{\bf Beyond Turing Machines\thanks{This work is licensed under the Creative Commons Attribution-No Derivative Works 3.0 Unported License (see http://creativecommons.org/licenses/by-nd/3.0/).}}
\author{Kurt Ammon \\ www.cstruct.org}
\date{}
\newcommand{\bc}{\begin{center}}
\newcommand{\ec}{\end{center}}
\newcommand{\bci}{\begin{center} \begin{minipage}{11.6cm} }
\newcommand{\eci}{\end{minipage} \end{center} }
\newcommand{\be}{\begin{enumerate}}
\newcommand{\ee}{\end{enumerate}}
\newcommand{\beq}{\begin{equation}}
\newcommand{\eeq}{\end{equation}}  
\newcommand{\ii}{\item}
\begin{document}
\maketitle

\begin{abstract}
{\it This paper discusses "computational" systems
capable of "computing"     functions not computable by predefined Turing machines
if the systems are not isolated from their environment.
Roughly speaking, these systems can change their finite descriptions      
by interacting with their environment.}
\end{abstract}

\section{Introduction}                       
\noindent
Turing \cite{Turing36} introduced the concept of "computing machines"
which subsequently were called Turing machines.
He proved that Hilbert's decision problem (Entscheidungs\-problem) is unsolvable,
that is,
there is no Turing machine determining whether 
or not a given statement in first-order 
predicate calculus (a mathematical proposition in number theory)
can be proved.
Wegner \cite{Wegner97} %
writes that Turing's 
precise characterization
of what can be computed established the respectability 
of computer science as a discipline.
He argues 
that Turing machines cannot capture the intuitive notion of what computers 
compute when computing is extended to include interaction.
His interaction machines have been criticized 
as an unnecessary Kuhnian paradigm shift \cite{WegnerGoldin03}. %
Prasse and Rittgen \cite{PrasseRittgen98} write
that Wegner's "interaction machines cannot compute non-recursive
functions, so Church's thesis still holds".
This implies that interaction machines
cannot "compute" functions not computable by 
Turing machines.        

This paper proves that there is no Turing machine 
producing a sequence of all computable functions on the
set of natural numbers. The proof implies the 
existence of "computational" systems
that cannot be modeled by 
predefined Turing machines
if the systems are not isolated from their environment.
Roughly speaking, these systems
change their finite descriptions      
by interacting with an environment
whose development
is not completely predictable for practical and theoretical reasons.
I argue that the proof even applies to existing
systems such as the Internet.
Finally, I introduce a new type of systems by
requiring that they be capable of "computing"
functions not computable by Turing machines. 
 
\section{Incompleteness Theorem}                       

A {\em computable function} is a function that can be computed by a Turing 
machine, that is, it can be represented by an ordinary computer program. Thus, a computable 
function on the set of natural numbers {\bf N} into {\bf N} can be regarded as a computer program 
producing a natural number in its output from any natural number in its input. An example of 
such a function $f$ is $f(n)=n+1$ which produces the successor $n+1$
of any natural number $n$.                                       

The computable functions
on the set of natural numbers %
can be regarded as models of computer 
programs and systems.                  
The restriction to 
computable functions
on the set of natural numbers %
is not relevant because
inputs and outputs 
of computer programs and systems
are represented as binary digits.
Useful 
programs and systems
should work for defined sets    
of inputs which correspond to subsets
of the set of natural numbers.
Such a definition can be extended to all natural numbers by assuming
a default output for inputs for which the program or
system is not defined. %
Ordinarily, a program or system is defined on a decidable 
set of inputs, that is, there is a program deciding whether
or not an input is admissible.
Thus, the computable functions on the set of natural numbers
model an important class of 
computer programs and systems.

{\bf Incompleteness Theorem:} There is no Turing machine producing a 
sequence of all computable functions $f_1$, $f_2$, ... on 
the set of natural numbers {\bf N} into {\bf N}.
 
{\bf Proof}. We assume that there is such a Turing machine. 
We define a new computable function 
$g$ on the  set of  natural numbers by  $g(n)=f_n(n)+1$ for all natural numbers $n$. Because the 
sequence $f_1$, $f_2$, ... contains all computable functions according to our original assumption, 
there is a natural number $n$ such that $g(x)=f_n(x)$ for all natural numbers $x$. This immediately 
yields the contradiction $g(n)=f_n(n)$ and $g(n)=f_n(n)+1$ because of the definition of the 
function 
$g$. This means that our original assumption is false, that is, there is no Turing machine 
producing all computable functions $f_1$, $f_2$, ... on the  set of  natural numbers.
 
The construction of the computable function $g$
in the proof can be represented by a Turing 
machine $T$. Of course, $T$ can be incorporated into any Turing machine. The proof implies that 
no Turing machine can produce all computable functions $f_1$, $f_2$, ... on the set of natural 
numbers no matter how $T$ is incorporated. In particular, $T$ can be incorporated into the Turing 
machine whose existence is assumed in the proof so that the extended Turing machine 
produces the sequence $f_1$, $f_2$, ... and  the function $g$. 
But the extended machine is different 
from the original machine and thus its application would contradict our assumption that there 
is a (single) Turing machine producing all computable functions on the set of natural 
numbers. 

Because any formal system can simply be defined as a theorem-proving Turing machine (see, 
for example, \cite[p.\ 72]{Goedel64}), 
the theorem also implies that any formal  theory is 
incomplete in the sense that it cannot "capture" all computable functions on the set of natural 
numbers. 

\section{Creative Systems}

Let $C$ be a system capable of performing the reasoning 
processes required for proving 
my 
simple incompleteness theorem. Thus, $C$ is capable of  constructing a 
computable function $g$ on the set of natural numbers not 
produced by any given Turing 
machine $M$. Figure~\ref{fig} illustrates the corresponding proof
which shows
that there is no predefined Turing machine $M$
producing all computable functions $f_1$, $f_2$, ...
on the set of 
natural numbers $C$ can construct. 

\begin{figure}
\setlength{\unitlength}{1cm}
\begin{center}
\begin{picture}(8,4)(-4,-2.2) 
\thicklines
\put(-6,.85){\Huge $C$}
\put(-5.2,1){\vector(1,0){3.5}}
\put(-5.7,.3){\vector(0,-1){2}}
\put(-1.4,.85){\Huge $M$}
\put(-.5,1){\vector(1,0){3.5}}
\put(3.4,.85){\Large              $f_1$, $f_2$, ...}
\put(-5.85,-2.45){\Huge  $g$ }
\put(-5.2,-2.45){\Large $\neq$        $f_1$, $f_2$, ...}
\put(-4,1.3){\Large\it uses}
\put(-5.1,-.7){\Large\it constructs}
\put(0,1.3){\Large\it produces}
\end{picture}
\end{center}
\caption{Proof\label{fig}}
\end{figure}                   

A difference between $C$ and a Turing machine is that $C$ is not regarded as a
static predefined object isolated from its environment. 
Rather, $C$ and the Turing machine $M$
in the proof are regarded as separate
interacting entities in the sense that $C$ 
assumes the existence of a Turing machine $M$
producing all computable functions $f_1$, $f_2$, ... 
on the set of natural numbers in order to
construct a computable function $g$ not produced by 
$M$. 
Roughly speaking, $C$ can
change itself by interacting with its environment, that is, the Turing machine $M$.

Because the capabilities of $C$ to construct computable functions cannot 
be modeled by 
any predefined Turing machine, 
another model of such systems seems to be required.

Computable functions can be represented by Turing machines or computer programs, that is,
they have finite descriptions. The set 
of such descriptions can be 
effectively enumerated.
This means that there is a Turing machine generating
a sequence containing all finite descriptions,
for example, in ascending length. 
For  these reasons, my simple incompleteness theorem implies 
that the function deciding whether or not
a given description represents a computable function on the set of natural numbers is not 
computable. If this function were computable, its application to an effective enumeration of 
descriptions would yield an effective enumeration of all computable functions on the set of 
natural numbers. This contradicts my theorem. 
Thus, there is no 
Turing machine 
deciding whether or not a given description
is a computable function, that is,
this decision problem (Entscheidungsproblem) is unsolvable.

Furthermore, my simple incompleteness theorem implies that systems capable of 
proving this theorem
seem to be capable of "solving"
the above decision problem {\it in the sense}
that they can construct more and more powerful computable functions on the set of 
natural numbers beyond the limits of any predefined Turing machine. 
This suggests the following definition 
of a new type of systems:
\bci
{\bf Definition:} A system is called {\it creative} if it is capable of 
"computing" non-computable 
functions, that is, determining values of non-computable functions for given arguments.
\eci 
With regard to this definition, creative systems can contain any computable function. Thus, 
they may be regarded as an extension of the concept of Turing machines in the sense that they 
can "compute" computable and non-computable functions.

An architecture of creative systems was
developed on the basis of experiments with the SHUNYATA 
program \cite{Ammon88,Ammon93}.
It is the first step towards the implementation of
a creative system. 
Roughly speaking, a creative system comprises
a reflection base 
and a varying number of
evolving analytical spaces.
\bci
{\bf Definition:} The {\em reflection base} contains a 
universal programming language, elementary domain-specific
concepts, and knowledge about this language and these concepts.
\eci
Because all computable 
functions can be represented in a universal programming language,
my simple incompleteness theorem implies 
that the reflection base cannot be formalized completely.
\bci
{\bf Definition:} {\em Analytical spaces} contain
partial knowledge whose domains of application are limited but
ordinarily expand in the course of the development
of the analytical spaces.
\eci
Roughly speaking, the development of new knowledge in creative systems
can be summarized in the following principles:
\bci
{\bf Principles of Development:} \em 
\be
\ii The knowledge in analytical spaces
arises from the reflection base and preceding 
knowledge.
\ii The development of knowledge involves the 
generation of new analytical spaces and the unification of existing %
ones. 
\ii
The economical variations of new 
knowledge tend to be preserved and the uneconomical ones to be destroyed. %
\ee
\eci
For example, the reflection base may contain
elementary knowledge about constants and functions of
a programming language.

A very simple programming task is the construction of a program
producing the successor $n$+1 of any natural number $n$ in its input.
This program can be constructed
on the basis
of the elementary knowledge that 1 is a natural number
and $x+y$ is a natural number for any
natural numbers $x$ and $y$.

Another simple example is the construction of a
Quicksort program  \linebreak[4] ${\it sort} (L)$,       
which sorts the elements of a list $L$ according to a given order relation 
"$x<y$" ($x$ is less than $y$) between any elements $x$ and $y$ of $L$.
The core of such a program can be represented by the 
functional pseudocode %
\begin{eqnarray}
  & \lefteqn {{\it append} ( {\it sort} (x \in L: x < {\it first} (L) ),} \hspace*{1.2cm} & \nonumber \\
 & & {\it first} (L), \nonumber \\
 & &  {\it sort} ( x \in L: {\it first} (L) < x) ) ),     \label{qs}
\end{eqnarray}                                              
where the {\it append} function appends lists and 
${\it first} (L)$ is the first element of a list $L$.
In order to sort a list $L$, the program (\ref{qs})
sorts the elements $x \in L$ that are less than its first element
${\it first} (L)$ and 
the elements $x \in L$ that are greater than 
${\it first} (L)$.
The recursive application of this 
"divide-and-conquer" strategy yields smaller and smaller partial lists
or empty lists which need not be sorted.        
Finally, 
the program (\ref{qs}) generates a sorted
list containing all elements of $L$ by successively appending all partial lists
previously sorted.

The Quicksort program (\ref{qs}) can also be constructed
on the basis of elementary knowledge 
about the functions
it contains.
Roughly speaking,
this knowledge need merely give the 
domains 
and ranges
of the functions, that is,
the sets on which the functions are defined
and the sets of values the functions may take on.
For example,    
the proposition $x < {\it first} (L)$
in (\ref{qs})
can be constructed on the basis of the knowledge
that ${\it first} (L)$ is an element of $L$
and $x < y$ is a proposition for
any elements $x \in L$ and $y \in L$.
An efficient selection of the functions 
used in 
the Quicksort program (\ref{qs}) seems feasible
because they 
are very elementary such as the 
{\it append} function
or even explicitly contained 
in the programming task such as
the order relation "$<$". 

For example, the SHUNYATA program 
generated theorem-proving programs by analyzing proofs of 
simple theorems on the basis of elementary
knowledge about the functions
the programs are composed of
\cite{Ammon88}. 
The development of these programs
illustrates  
very simple aspects of the principles given above,
for example,
the unification of analytical spaces
and the generation of economical variations.
The theorem-proving programs developed by SHUNYATA
generated proofs of 
further theorems, in particular, 
a proof of SAM's Lemma which is simpler than any proof 
known before. 
The complexity of SAM's Lemma more or less %
represented the state of the art 
in automated theorem proving \cite[p.\ 561]{Ammon88}.

G\"odel's incompleteness theorem
says that every formal number theory contains an undecidable
formula, that is, neither the formula nor its negation are
provable in the theory.
The main problem in the proof of G\"odel's theorem is the
construction of such a formula.
Analogously to the construction of
the Quicksort program (\ref{qs}),
an undecidable formula can be constructed
on the basis
of elementary rules for the formation
of formulas, that is, 
on the basis of elementary knowledge
about the symbols the formula 
contains.
Ammon \cite{Ammon93} describes
a proof of G\"odel's theorem and refers to further
experiments with the SHUNYATA program.

\section{Discussion}                       

The Internet is a network of a varying number of computer programs and systems.
My simple incompleteness theorem implies that
no predefined formal system 
can completely model
such a network
because a program (computable function)
not "captured" by the formal system
can be constructed and added to the network.
The proof of my theorem implies that the construction of this
program can be achieved automatically. 

Systems capable of communicating and interacting
with humans more naturally
than existing systems
should be capable of reconstructing 
computable functions
implicitly used in human forms of communication.
My theorem implies that 
there is no predefined Turing machine or formal system 
modeling this communication.

Computer programs must be 
tested and debugged before they can be used in practice.
My theorem implies
that there is no general predefined algorithm
for the verification of programs
because a program (computable function) not "captured" by the algorithm
can be constructed from the algorithm itself.
Rather, the verification of programs is achieved 
on the basis of experience.

The Turing machine 
producing a 
sequence of computable functions 
$f_1$, $f_2$, ...
in the proof of my incompleteness theorem
can be 
regarded as an analytical space.
The construction of 
another Turing machine
producing        
$f_1$, $f_2$, ... and
a computable function $g$ not produced by the original machine
can be regarded as 
the construction of a new analytical space.
Thus, my simple theorem implies 
that all analytical spaces cannot be unified into a single analytical
space. Roughly speaking, the development of new knowledge in analytical spaces
cannot be regarded as a completely describable "closed box".
Rather, it is an open process which may transcend any frame
specified in advance.

My work can also be regarded as an investigation 
with the aid of computers why Hilbert's decision 
problem 
(Entscheidungsproblem)
is unsolvable because creative systems 
can determine 
 beyond the limits of any predefined algorithm
whether 
or not a statement in predicate calculus 
can be proved.

Church's thesis                                 
states that every effectively calculable function 
is general recursive \cite[pp.\ 317-323]{Kleene52}.
Turing's thesis, which is equivalent to Church's thesis,
states that every function
that would be naturally regarded as computable
is computable under his definition, that is, by a 
Turing machine \cite[pp.\ 376-381]{Kleene52}.
My work should not be regarded as a refutation of 
Church's or Turing's thesis.
Rather, it sheds new light on these theses 
and on Hilbert's decision problem (Entscheidungsproblem).
In particular, it proves the existence of systems "computing"
non-computable functions if "computing" means
determining values of functions for given arguments.
But these systems have no complete 
finite descriptions
that can be given in advance. Rather, they 
can transcend any predefined formal description.

\section{Related Work}    
Turing \cite{Turing50,Turing69}
discusses whether "it is possible for machinery
to show intelligent behaviour".
Referring to G\"odel's theorem and
other, in some respects similar, 
results due to Church, Kleene, Rosser, and himself,
Turing \cite[p. 445]{Turing50} writes
"that there are limitations to the powers of
any particular machine".
Turing \cite[p.\ 4]{Turing69}     
states:
\bci
The argument from G\"odel's and other theorems ... rests essentially on the condition 
that the machine must not make errors. But this is not a requirement for intelligence.
\eci
He %
argues that a "man 
provided with paper, pencil, and rubber, 
and subject to strict discipline" 
can "produce the effect of a computing 
machine", 
but             %
"discipline is certainly not enough in itself to produce intelligence"
\cite[p.\ 9 and p.\ 21]{Turing69}.
My simple incompleteness theorem confirms 
that the powers of any particular Turing machine are limited.
The theorem implies that
systems capable of proving the theorem and interacting with
their environment cannot be modeled by any 
predefined Turing machine. 
Thus, "ordinary computational systems"
suitably equipped 
to prove the theorem and to interact with
their environment
can in principle transcend
the powers of any machine completely specified in advance.
My theorem implies that such systems are necessarily 
empirical and fallible in the sense
that a complete predefined formalization of 
their truth judgments, for example, whether a given
computer program represents a computable function
on the set of natural numbers, is impossible.

My theorem and principles about creative systems are
compatible with Post's view \cite[p.\ 417]{Post65}:
\bci
... logic must ... in its very operation be informal. Better still, we write
\bc\vspace{-.2cm}
                 The Logical Process is Essentially Creative.
\ec
\eci

Wegner \cite{Wegner97} 
argues that "interaction is more powerful than algorithms".
My simple incompleteness theorem can be regarded
as a mathematical proof             
of his thesis.
Moreover, the theorem and its proof imply the existence of systems
"computing" non-computable functions,
that is, determining values of non-computable functions
for given arguments.
In this sense, they confirm the view of
Wegner and Goldin \cite{WegnerGoldin03}
"that neither logic nor algorithms can completely model computing and 
human thought."            

Wegner \cite{Wegner97} 
writes that the incompleteness of interaction machines %
follows from the fact that dynamically 
generated input streams are 
mathematically modeled by infinite sequences
over a finite alphabet, which are not enumerable.
The "incompleteness", rather indeterminacy of creative systems follows from
their definition, in particular from
the fact
that no formal theory 
can "capture" all computable functions on the set of natural 
numbers.

My incompleteness theorem implies
that there is no general algorithm or formal system
for the verification of programs.
This result is compatible with Wegner's 
view 
"that proving correctness is not merely hard but impossible"
because
"{\em open, empirical, falsifiable, or interactive systems 
are necessarily incomplete}" \cite[p.\ 10]{Wegner97}.
It is also compatible with
G\"odel's statement \cite[p.\ 84]{Goedel46}
that 
"one has been able to define them [demonstrability and definability] only relative 
to a given language", that is, there is no general 
definition of formal proofs but such definitions
can only be given in particular formal systems.

\end{document}